\documentclass[conference]{IEEEtran}

\PassOptionsToPackage{hyphens}{url}
\usepackage{booktabs}
\usepackage{listings}
\usepackage{tikz}
\usepackage{hyperref}
\usepackage{xcolor}
\usetikzlibrary{positioning, arrows.meta, shapes.geometric, fit, backgrounds, shadows}

\definecolor{codebg}{HTML}{F7F7F7}
\definecolor{codeframe}{HTML}{CCCCCC}
\definecolor{kwcolor}{HTML}{0033B3}
\definecolor{strcolor}{HTML}{067D17}
\definecolor{cmtcolor}{HTML}{8C8C8C}
\definecolor{numcolor}{HTML}{1750EB}

\lstdefinelanguage{TypeScript}{
  keywords={import,export,from,const,let,var,function,async,await,return,
             interface,type,class,extends,implements,new,if,else,throw,try,
             catch,for,of,in,true,false,null,undefined,string,number,boolean,
             void,Promise,Record,Array,Partial,Omit,Pick,default},
  keywordstyle=\color{kwcolor}\bfseries,
  stringstyle=\color{strcolor},
  commentstyle=\color{cmtcolor}\itshape,
  numberstyle=\color{numcolor},
  morestring=[b]',
  morestring=[b]",
  morestring=[b]`,
  morecomment=[l]{//},
  morecomment=[s]{/*}{*/},
  sensitive=true
}

\lstdefinelanguage{YAML}{
  keywords={true,false,null},
  keywordstyle=\color{kwcolor},
  stringstyle=\color{strcolor},
  commentstyle=\color{cmtcolor}\itshape,
  morecomment=[l]{\#},
  sensitive=true
}

\lstdefinelanguage{SQL}{
  keywords={CREATE,TABLE,PRIMARY,KEY,DEFAULT,NOT,NULL,UNIQUE,CHECK,IN,
             REFERENCES,GENERATED,ALWAYS,AS,STORED,boolean,integer,text,
             jsonb,uuid},
  keywordstyle=\color{kwcolor}\bfseries,
  stringstyle=\color{strcolor},
  commentstyle=\color{cmtcolor}\itshape,
  morecomment=[l]{--},
  morecomment=[s]{/*}{*/},
  morestring=[b]',
  sensitive=false
}

\title{Skilldex: A Package Manager and Registry for Agent Skill Packages with Hierarchical Scope-Based Distribution}

\author{
  \IEEEauthorblockN{Sampriti Saha, Pranav Hemanth}
  \IEEEauthorblockA{
    Pandemonium Research\\
    \{sampritisaha, pranavhemanth\}@pandemoniumresearch.com
  }
}

\begin{document}

\maketitle

\begin{abstract}
Large Language Model (LLM) agents are increasingly extended at runtime via \emph{skill packages}, structured natural-language instruction bundles loaded from a well-known directory. Community install tooling and registries exist, but two gaps persist: no public tool scores skill packages against Anthropic's published format specification, and no mechanism bundles related skills with the shared context they need to remain mutually coherent.

We present \textbf{Skilldex}, a package manager and registry for agent skill packages addressing both gaps. The two novel contributions are: (1) \textbf{compiler-style format conformance scoring} against Anthropic's skill specification, producing line-level diagnostics on description specificity, frontmatter validity, and structural adherence; and (2) the \textbf{skillset abstraction}, a bundled collection of related skills with shared assets (vocabulary files, templates, reference documents) that enforce cross-skill behavioral coherence --- a property independently installed skills cannot guarantee. Skilldex also provides supporting infrastructure: a three-tier hierarchical scope system (global, shared, and project), a human-in-the-loop agent suggestion loop, a metadata-only community registry, and a Model Context Protocol (MCP) server exposing all operations to agents natively. The system is implemented as a TypeScript CLI (\textit{skillpm} / \textit{spm}) with a Hono/Supabase registry backend, and is open-source.
\end{abstract}


\section{Introduction}
\label{sec:intro}

LLM-powered agents are increasingly extended via \emph{skills}: structured natural-language documents that instruct an agent to adopt a specific behavior, follow a particular methodology, or leverage a defined set of tools. Popularized by Anthropic's Claude Code platform and now adopted across a range of LLM agents, skills are formalized as \texttt{SKILL.md} files (Markdown with YAML frontmatter) that agents load from a well-known directory and use as prompting context during task execution~\cite{anthropic2024skills}.

Skill packages are analogous to software libraries: they encapsulate reusable functionality, have authors and versions, vary in quality, and are most valuable when discoverable and shareable. Install tooling and community registries~\cite{vercellabsskills} now exist for skills, but two gaps remain:

\begin{enumerate}
  \item \textbf{No spec-grounded conformance signal.} Existing tools surface install counts and publisher badges, but none scores a skill against Anthropic's published format specification. A skill whose description is too short to trigger reliably, or whose frontmatter is malformed, is published and installed identically to a well-formed one.

  \item \textbf{No skillset coherence mechanism.} Skills are installed individually as flat units. There is no abstraction for bundling related skills with the shared convention files and reference documents they need to remain mutually coherent --- installing related skills independently leaves each with its own implicit vocabulary that can silently diverge from the others.
\end{enumerate}

We present \textbf{Skilldex} (a portmanteau of \emph{skill} and \emph{index}, in reference to Anthropic's skills directory), a package manager and registry that closes both gaps while extending the skills directory with full package management semantics: hierarchical scoping, agent-driven suggestion, and a community registry. The remainder of this section enumerates Skilldex's two novel contributions and the supporting infrastructure that makes them practical.

\textbf{Novel contributions:}
\begin{itemize}
  \item A \textbf{format conformance scoring system} (Section~\ref{sec:validation}) producing a 0--100 score against Anthropic's skill format specification, with compiler-style line-level diagnostics.

  \item The \textbf{skillset abstraction} (Section~\ref{sec:skillsets}): a bundled, installable unit of related skills with shared assets (convention files, templates, persona definitions). Skills in the same skillset reference the same files, keeping their outputs mutually coherent.
\end{itemize}

\textbf{Supporting infrastructure:}
\begin{itemize}
  \item A \textbf{hierarchical scope system} (Section~\ref{sec:scope}) with three levels (global, shared, project) and local-first precedence, keeping loaded context proportional to the current task.

  \item A \textbf{human-in-the-loop suggestion loop} (Section~\ref{sec:suggestion}) that surfaces a proposed skill manifest for review before an agent begins work.

  \item A \textbf{metadata-only community registry} (Section~\ref{sec:registry}) with conformance scores in skill metadata, full-text search, and a trust tier model seeded with Anthropic's official skills.
\end{itemize}

\section{Background and Related Work}
\label{sec:background}

\subsection{Agent Skill Packages}

The \texttt{SKILL.md} format was introduced by Anthropic's Claude Code platform and has since been adopted across a growing range of LLM agents from multiple providers~\cite{anthropic2024skills,vercellabsskills}. A skill is a directory containing a \texttt{SKILL.md} file with a standardized YAML frontmatter block followed by Markdown content:

\begin{lstlisting}[language=YAML, caption={Example SKILL.md frontmatter}]
---
name: forensics-memory-analysis
description: "Guides Claude through systematic memory dump
  analysis using Volatility3 and similar tools. Covers
  process enumeration, network connections, and artifact
  extraction for incident response."
version: "1.0.0"
tags: [forensics, memory, volatility, incident-response]
author: "skilly"
spec_version: "1.0"
---
## Instructions
...
\end{lstlisting}

The skill directory may additionally contain a \texttt{scripts/} subdirectory for executable helpers, a \texttt{references/} subdirectory for supplementary documentation, and an \texttt{assets/} directory for images and data files. In Claude Code, skills are loaded from \texttt{.claude/skills/} at project scope and from \texttt{\textasciitilde/.claude/skills/} at user scope; other agents use analogous well-known directories.

\subsection{Package Management Systems}

Package management is a solved problem in software engineering. npm~\cite{npm}, pip~\cite{pip}, cargo~\cite{cargo}, and Homebrew~\cite{homebrew} all implement variants of the same core operations: install, remove, update, search, and publish. The design space centers on a few key decisions: binary vs.\ source distribution, centralized vs.\ decentralized registry, flat vs.\ hierarchical dependency resolution. Skilldex draws most directly from npm and pip in its CLI design, and from Python virtual environments in its scoping model.

\subsection{MCP and Agent Tool Extension}

The Model Context Protocol (MCP)~\cite{mcp2024} provides a standardized interface for exposing tools and resources to LLM agents. Skilldex implements an MCP server that exposes all core operations (\texttt{skilldex\_install}, \texttt{skilldex\_list}, \texttt{skilldex\_validate}, \texttt{skilldex\_search}, \texttt{skilldex\_suggest}, \texttt{skilldex\_uninstall}) as callable tools, enabling agents to manage their own skill environment mid-session without user terminal interaction.

\subsection{Related Systems}

\textbf{Smithery.ai} and \textbf{Glama.ai} are registries for MCP servers, a different layer of abstraction (MCP servers expose tools; agent skills extend behavior).

\textbf{LangChain Hub}~\cite{langchainhub} provides a registry for LangChain prompts and chains, targeting a different ecosystem and abstraction level.

\textbf{Anthropic's public skills directory}~\cite{anthropic2024skills} is a structured GitHub repository hosting official skills, the Agent Skills specification, and a plugin marketplace. It has no CLI install tooling and no scoping model. Skilldex treats it as a seeding source for the \texttt{verified} trust tier.

\textbf{vercel-labs/skills}~\cite{vercellabsskills} is an open agent skills CLI (\texttt{npx skillsadd}) and community registry (skills.sh) that installs \texttt{SKILL.md}-based packages across 40+ agents, with search, install counts, and official publisher designations. Skilldex differs in offering hierarchical scoped installation and spec-grounded format conformance scoring, plus the skillset abstraction.

Other community efforts (e.g., CCPM, ClawdHub, claude-skill-registry) provide skill installation, hash-based verification, or basic validation, but none publishes a scoring rubric grounded in Anthropic's \texttt{SKILL.md} specification or supplies the skillset bundling abstraction described here.

\section{System Architecture}
\label{sec:architecture}

\begin{figure*}[t]
\centering
\begin{tikzpicture}[
  font=\small,
  node distance=0.8cm and 1.2cm,
  box/.style={draw, rounded corners=3pt, minimum width=2.8cm, minimum height=0.65cm,
              align=center, fill=white, drop shadow},
  registry/.style={box, fill=blue!8},
  core/.style={box, fill=green!8},
  interface/.style={box, fill=orange!8},
  arrow/.style={-Stealth, thick},
  label/.style={font=\footnotesize\itshape, text=gray}
]

\node[interface] (cli) {CLI\\\texttt{skillpm / spm}};
\node[interface, right=1.4cm of cli] (mcp) {MCP Server\\\texttt{stdio}};

\node[core, below=1.2cm of cli] (installer) {Installer};
\node[core, right=0.6cm of installer] (resolver) {Scope Resolver};
\node[core, right=0.6cm of resolver] (validator) {Validator};

\node[core, below=0.8cm of installer] (manifest) {Manifest I/O};
\node[core, below=0.8cm of validator] (suggest) {Suggest Agent};

\node[registry, right=1.2cm of mcp] (regclient) {Registry Client};
\node[registry, right=1.2cm of regclient] (regapi) {Registry API\\\textit{(Hono/Vercel)}};
\node[registry, below=0.8cm of regapi] (db) {Supabase\\\textit{(PostgreSQL)}};
\node[registry, below=0.8cm of db] (ghfetch) {GitHub Fetch};

\draw[arrow] (cli.south) -- (installer.north);
\draw[arrow] (cli.south east) -- (resolver.north west);
\draw[arrow] (cli.south east) -- (validator.north west);

\draw[arrow] (mcp.south west) -- (installer.north east);
\draw[arrow] (mcp.south) -- (resolver.north east);
\draw[arrow] (mcp.south) -- (validator.north);

\draw[arrow] (installer.south) -- (manifest.north);
\draw[arrow] (installer.east) -- (resolver.west); 

\draw[arrow] (mcp.south east) to[out=-45, in=45] (suggest.east);

\draw[arrow] (cli.north east) to[out=20, in=160] (regclient.north west);
\draw[arrow] (mcp.east) -- (regclient.west);

\draw[arrow] (regclient.east) -- (regapi.west);
\draw[arrow] (regapi.south) -- (db.north);
\draw[arrow] (db.south) -- (ghfetch.north);

\node[label, below=1cm of current bounding box.south, anchor=center] {CLI / MCP: interface layer \quad Core: local logic \quad Registry: network layer};

\end{tikzpicture}
\caption{Skilldex system architecture. The CLI and MCP server share all core modules. The registry is a separate service accessed via a typed HTTP client.}
\label{fig:architecture}
\end{figure*}

Skilldex is structured as three independent components:

\begin{enumerate}
  \item \textbf{Skilldex CLI} (\texttt{skilldex-cli}): a Node.js 20+ npm package providing the \texttt{skillpm} and \texttt{spm} commands. Implemented in TypeScript with Commander, simple-git, and Zod.

  \item \textbf{Skilldex Registry} (\texttt{skilldex-registry}): a Hono web application on Vercel, backed by Supabase (PostgreSQL). Stores skill and skillset metadata; handles auth, search, and install-count tracking.

  \item \textbf{Skilldex Web}: a Next.js application at \texttt{skilldex-web.vercel.app} providing the registry browser UI and documentation site.
\end{enumerate}

The CLI and MCP server share the same \texttt{core/} modules; both interfaces invoke the same install, validate, resolve, and manifest functions. Fig.~\ref{fig:architecture} shows the full system.

\section{Hierarchical Scope System}
\label{sec:scope}

\subsection{Motivation}

Loading every installed skill into every agent session is the simplest approach but fails at scale. Context window consumption is proportional to the number of skills loaded; skill descriptions consume tokens even when irrelevant; and name collisions between a global default skill and a project-specific override have no principled resolution strategy.

The Python virtual environment model~\cite{virtualenv} and CSS cascade~\cite{csscascade} both solve structurally analogous problems: how does a consumer resolve the same name from multiple potential sources, and which one wins? Skilldex adapts this pattern for skill installation.

\subsection{Three-Tier Hierarchy}

Skills are installed at one of three scope levels:

\begin{itemize}
  \item \texttt{global}: Available to all projects. Stored at \texttt{\textasciitilde/.skilldex/global/}. Intended for universal capabilities such as writing style guides, general debugging methodologies, and documentation standards.

  \item \texttt{shared}: Available across multiple explicitly opted-in projects. Stored at \texttt{\textasciitilde/.skilldex/shared/}. Intended for cross-project team conventions that should not pollute every project's scope.

  \item \texttt{project}: Scoped to a single project. Stored at \texttt{<project-root>/\allowbreak.skilldex/}. The project root is located by walking parent directories for a \texttt{.git} directory or \texttt{package.json}, falling back to \texttt{cwd}.
\end{itemize}

Each scope level maintains its own \texttt{skilldex.json} manifest and \texttt{skills/} directory. The manifest tracks installed skills, their source URLs, scores, and installation timestamps:

\begin{lstlisting}[language=TypeScript, caption={Manifest schema (Zod)}]
const SkillManifestSchema = z.object({
  skilldexVersion: z.string(),
  scope: z.enum(['global', 'shared', 'project']),
  skills: z.record(InstalledSkillSchema),
  skillsets: z.record(InstalledSkillsetSchema).default({}),
  updatedAt: z.string(),
})
\end{lstlisting}

\subsection{Resolution Rules}

Resolution follows a \emph{local-first precedence} rule: lower scope always overrides higher scope for the same skill name. A \texttt{project}-scope installation of \texttt{forensics-memory-analysis} shadows a \texttt{global}-scope installation of the same name. This allows project-specific customization without modifying shared or global installations.

The scope resolver maps a \texttt{ScopeLevel} to a concrete \texttt{ScopeConfig} containing the root path, manifest path, skills directory, and skillsets directory:

\begin{lstlisting}[language=TypeScript, caption={ScopeConfig interface}]
interface ScopeConfig {
  level: ScopeLevel     // 'global' | 'shared' | 'project'
  rootPath: string
  manifestPath: string  // <root>/skilldex.json
  skillsDir: string     // <root>/skills/
  skillsetsDir: string  // <root>/skillsets/
}
\end{lstlisting}

The \texttt{resolveAllScopes()} function materializes all three \texttt{ScopeConfig} objects simultaneously, which the installer uses for cross-scope conflict detection.

\subsection{Installation Sources}

The install command accepts three source forms:

\begin{enumerate}
  \item \textbf{Registry name}: \texttt{skillpm install forensics-memory-analysis}. Looks up the skill in the registry, retrieves the source URL, and delegates to the git path.
  \item \textbf{Git URL}: \texttt{skillpm install} followed by a \url{git+https://}~URL such as \url{git+https://github.com/user/repo/tree/main/skills/my-skill}. Supports branch and subpath syntax. Clones to a temp directory, discovers skill directories, installs each.
  \item \textbf{Local path}: \texttt{skillpm install ./my-skill}. Validates and copies directly.
\end{enumerate}

All three paths converge on \texttt{installFromPath(sourcePath, options)}, which performs validation, copies the skill directory into the target scope, and updates the manifest atomically.

\subsection{Cross-Scope Conflict Detection}

Before installation, the installer reads all three scope manifests and emits a warning if the skill is already installed at a different scope. This is informational and does not block installation, consistent with Skilldex's general principle that warnings never gatekeep.

\section{Format Conformance Scoring}
\label{sec:validation}

\subsection{Motivation}

Undertriggering, where an agent fails to invoke a skill when it should, is documented as a known failure mode in Anthropic's skill creator guide~\cite{anthropic2024skills}. The primary cause is description quality: if a skill's description is too short, too generic, or poorly worded, the agent's context-based skill selection will not surface it at the right moment.

Format conformance scoring provides a measurable, objective proxy for one of the most impactful levers a publisher has: description specificity and length. It is explicitly \emph{not} a measure of functional quality: a syntactically perfect skill can be useless, and a low-scoring skill can be genuinely valuable. This disclaimer appears prominently wherever a score is displayed.

\subsection{Scoring Checks}

The validator performs eight checks for skills, producing a score from 0 to 100:

\begin{table*}[t]
\centering
\caption{Skill format conformance checks}
\label{tab:skill-checks}
\begin{tabular}{lrp{9cm}}
\toprule
\textbf{Check} & \textbf{Pts} & \textbf{Rationale} \\
\midrule
YAML frontmatter parseable & 25 & Fatal if missing; no further checks run \\
\texttt{name} field present & 10 & Required for registry and manifest \\
\texttt{description} present & 10 & Primary triggering mechanism \\
Description $\geq 30$ words & 10 & Specificity threshold \\
\texttt{SKILL.md} $\leq 500$ lines & 15 & Token budget constraint \\
Allowed subdirectories only & 10 & Enforces \texttt{scripts/}, \texttt{references/}, \texttt{assets/} \\
Referenced resources exist & 15 & No broken relative links \\
Resources in correct subdirs & 5 & Scripts in \texttt{scripts/}, docs in \texttt{references/} \\
\bottomrule
\end{tabular}
\end{table*}

Missing frontmatter is the only truly fatal condition: it scores zero and halts further checks, because all subsequent checks depend on parsed frontmatter. All other failures produce diagnostics but allow installation to proceed.

\subsection{Diagnostic Output}

Diagnostics follow a compiler-style format with severity (\texttt{error}, \texttt{warning}, \texttt{pass}), an optional line number, and a human-readable message:

\begin{lstlisting}[language={}, caption={Example validation output}]
  pass    YAML frontmatter valid
  pass    name field present
  error   line 4: description too short (12 words,
          recommended: 30+)
  pass    SKILL.md line count OK (87 lines)
  warning Unknown subdirectory "helpers" -- only
          scripts/, references/, assets/ allowed
  pass    All referenced resources exist

Format conformance score: 45/100
Validated against: skill-format v1.0
\end{lstlisting}

The \texttt{--json} flag on all commands emits a structured JSON object for programmatic consumption, enabling CI/CD integration:

\begin{lstlisting}[language=TypeScript, caption={ValidationResult type}]
interface ValidationResult {
  skill: string
  score: number          // 0-100
  diagnostics: ValidationDiagnostic[]
  specVersion: string    // e.g. "1.0"
  passCount: number
  warnCount: number
  errorCount: number
}
\end{lstlisting}

\subsection{Spec Ownership Boundary}

A deliberate design decision is that Skilldex does not own or extend the skill format specification. The scoring table in Table~\ref{tab:skill-checks} is derived directly from Anthropic's published \texttt{SKILL.md} creator guide. If Anthropic updates the spec, Skilldex cuts a new scorer version (\texttt{spec\_version} in manifest) but does not make normative decisions about what the format should be. This boundary prevents Skilldex from becoming a de facto spec owner, a role it is not designed to hold.

\section{Agent Suggestion Loop}
\label{sec:suggestion}

\subsection{Motivation}

Most agent frameworks execute without a pre-task capability review. The agent loads whatever skills are installed and begins work. This conflates two decisions that should be separate: \emph{what capabilities should this agent have?} and \emph{how should it use them?}

Skilldex interposes an explicit human-in-the-loop checkpoint between project context and task execution. This maps to a broader principle in responsible AI deployment: explicit checkpoints before capability expansion, not after~\cite{seshia2016verified}.

\subsection{Workflow}

The \texttt{skillpm suggest} command (and equivalent \texttt{skilldex\_suggest} MCP tool) implements a three-phase workflow:

\textbf{Phase 1: Context Gathering.} The system reads the project's \texttt{README.md} (first 100 lines), \texttt{package.json} (name, description, scripts, dependencies), existing agent configuration directory contents, and installed skill manifests across all scopes. This is assembled into a structured context string.

\textbf{Phase 2: Proposal Generation.} The context string is passed to an LLM (via the Anthropic SDK) with a system prompt asking it to propose a skill manifest, a list of skills with names, justifications, and suggested scope levels. The prompt instructs the model to check whether proposed skills are already installed and to distinguish between skills that exist in the registry and custom skills that would need to be authored.

\begin{lstlisting}[language=TypeScript, caption={SuggestionProposal type}]
interface SuggestionProposal {
  skillName: string
  reason: string
  suggestedScope: ScopeLevel  // 'global'|'shared'|'project'
  available: boolean           // found in registry?
}
\end{lstlisting}

\textbf{Phase 3: Human Approval.} The CLI presents each proposal interactively, showing the skill name, reason, suggested scope, and registry availability. The user approves or rejects each proposal and optionally overrides the scope. Approved skills that are available in the registry are queued for manual installation via \texttt{skillpm install <name>}; the suggestion loop does not yet trigger installation directly. Approved skills that are not available in the registry are listed as authoring candidates.

\begin{lstlisting}[language={}, caption={Interactive suggestion output}]
Proposed skills for this project:
  forensics-memory-analysis  [community] [project scope]
  Reason: project contains Volatility scripts
  Approve? (Y/n/scope): y

  log-triage  [verified] [project scope]
  Reason: multiple log directories detected
  Approve? (Y/n/scope): y

  malware-static-analysis  [community] [global scope]
  Reason: project description mentions binary analysis
  Approve? (Y/n/scope): n

Approved 2 skill(s). To install, run:
  skillpm install forensics-memory-analysis --scope project
  skillpm install log-triage --scope project
\end{lstlisting}

The non-interactive \texttt{--yes} flag approves all proposals at their suggested scopes without prompting, intended for CI environments or automated setup scripts.

\section{Community Registry}
\label{sec:registry}

\subsection{Architecture Decisions}

The registry stores \emph{metadata only}. Skill files are not uploaded to or hosted by the registry. The \texttt{source\_url} field points to the GitHub repository (or subdirectory) where the skill lives, and installation fetches directly from GitHub. This decision has several consequences:

\begin{itemize}
  \item Infrastructure cost is near zero: no binary storage, no CDN.
  \item GitHub's existing reliability, access controls, and versioning are inherited.
  \item The author retains ownership and update authority over their skill content.
  \item The registry can be re-seeded from source without data loss.
\end{itemize}

The tradeoff is a dependency on GitHub availability and rate limits at install time. Skilldex optionally accepts a \texttt{GITHUB\_TOKEN} environment variable to use authenticated requests (5,000 req/hr) rather than anonymous ones (60 req/hr).

\subsection{Database Schema}

The registry backend uses PostgreSQL via Supabase. The core tables are:

\begin{lstlisting}[language=SQL, caption={Core schema (condensed)}]
CREATE TABLE publishers (
  id            uuid PRIMARY KEY DEFAULT gen_random_uuid(),
  github_handle text NOT NULL UNIQUE,
  verified      boolean DEFAULT false
);

CREATE TABLE skills (
  id            uuid PRIMARY KEY DEFAULT gen_random_uuid(),
  name          text NOT NULL UNIQUE,
  description   text NOT NULL,
  source_url    text NOT NULL,
  trust_tier    text NOT NULL
    CHECK (trust_tier IN ('verified', 'community')),
  score         integer CHECK (score BETWEEN 0 AND 100),
  spec_version  text NOT NULL,
  tags          text[],
  install_count integer DEFAULT 0,
  published_by  uuid REFERENCES publishers(id)
);

CREATE TABLE skillsets (
  -- mirrors skills with additional:
  skill_refs    jsonb NOT NULL DEFAULT '[]',
  skill_count   integer GENERATED ALWAYS AS
                (jsonb_array_length(skill_refs)) STORED
);
\end{lstlisting}

Full-text search is provided by PostgreSQL's native \texttt{to\_tsvector} over the name and description columns, with trigram indexes (\texttt{pg\_trgm}) for fuzzy matching on skill names.

\subsection{Trust Tier Model}

Skilldex uses a two-tier model:

\begin{itemize}
  \item \texttt{verified}: Reserved for Anthropic's officially published skills. Assignment is manual, by Skilldex maintainers. There is no automated promotion path.
  \item \texttt{community}: Default for all submissions. Any authenticated GitHub user can publish a community skill.
\end{itemize}

The binary model is a deliberate simplification. Star ratings, karma systems, and multi-tier promotion ladders all introduce inflation dynamics that erode the signal value of the tier over time~\cite{ratinginflation}. A hard boundary between Anthropic-official and community-published skills provides a clear, manipulation-resistant signal.

Importantly, trust tier, like validation score, never blocks installation. The registry surfaces information; it does not gatekeep. This respects user autonomy and avoids the adoption penalty that over-restrictive package managers incur.

\subsection{API Surface}

The registry exposes a versioned REST API at \texttt{/v1/}:

\begin{table*}[t]
\centering
\caption{Registry API endpoints}
\label{tab:api}
\begin{tabular}{llp{6cm}}
\toprule
\textbf{Method} & \textbf{Path} & \textbf{Purpose} \\
\midrule
GET    & \texttt{/skills}                  & Search/list with filters \\
GET    & \texttt{/skills/:name}            & Fetch single skill \\
GET    & \texttt{/skills/:name/install}    & Fetch install info + count \\
POST   & \texttt{/skills}                  & Publish new skill (auth) \\
PATCH  & \texttt{/skills/:name}            & Re-fetch and re-score (auth) \\
DELETE & \texttt{/skills/:name}            & Remove skill (auth) \\
\midrule
GET    & \texttt{/skillsets}               & Search/list skillsets \\
GET    & \texttt{/skillsets/:name}         & Fetch single skillset \\
GET    & \texttt{/skillsets/:name/install} & Fetch install info + count \\
POST   & \texttt{/skillsets}               & Publish new skillset (auth) \\
PATCH  & \texttt{/skillsets/:name}         & Re-fetch and re-score (auth) \\
DELETE & \texttt{/skillsets/:name}         & Remove skillset (auth) \\
\midrule
GET    & \texttt{/auth/github}             & Initiate GitHub OAuth \\
GET    & \texttt{/auth/github/callback}    & GitHub OAuth callback \\
GET    & \texttt{/auth/me}                 & Current publisher info \\
GET    & \texttt{/spec-versions}           & List spec versions \\
GET    & \texttt{/spec-versions/current}   & Current spec version \\
\bottomrule
\end{tabular}
\end{table*}

Authentication uses GitHub OAuth via Supabase Auth. Publishers are stored with their GitHub handle, which becomes the default \texttt{author} field for skills they publish. Rate limiting is applied per endpoint (100 req/min on search, 500 req/min on install); the current deployment uses in-memory rate limiting, with Upstash Redis planned for production hardening.

\subsection{Registry Seeding}

The \texttt{verified} tier is seeded from Anthropic's public skills directory via a two-stage process. First, source repositories are registered with the registry admin tooling (\texttt{npm run add-repo}), recording a GitHub repository URL and optional subdirectory subpath. Second, a nightly GitHub Actions workflow (scheduled at 02:00 UTC) fetches each registered source, validates the skill or skillset it finds, and upserts the result into the \texttt{skills} table with \texttt{trust\_tier = 'verified'}. The workflow can also be triggered manually via \texttt{npm run seed} (interactive) or \texttt{npm run seed:ci} (non-interactive, for scripted environments).

Because the registry stores metadata only, re-seeding is non-destructive: a source skill that has not changed produces an identical upsert. This means the nightly job also serves as a consistency check: if Anthropic updates a skill's description or version, the registry reflects the change within 24 hours without operator intervention.

\subsection{Publishing Flow}

Publishing a skill from the CLI requires a \texttt{SKILLDEX\_TOKEN} environment variable obtained from the registry's GitHub OAuth flow. The publish command: (1) reads the skill name from \texttt{SKILL.md} frontmatter; (2) detects the GitHub remote URL via \texttt{git remote get-url origin}; (3) normalizes SSH URLs to HTTPS; (4) \texttt{POST /v1/skills} with \texttt{\{name, source\_url, tags\}}. The registry backend then fetches and validates the skill from GitHub server-side, storing the result.

\section{Skillset Bundling}
\label{sec:skillsets}

\subsection{Motivation}

Agent use-cases are rarely defined by a single skill, and installing related skills independently produces a subtler problem than inconvenience: \emph{behavioral drift}. Consider a developer agent with two separate skills, one writing Conventional Commits messages and one generating CHANGELOG entries from git history. Installed independently, they share no understanding of how commit types map to changelog sections. If one recognizes a custom commit type the other does not, the workflow breaks --- not from a bug, but from a gap in shared vocabulary.

The \emph{skillset} is a publishable, installable unit that packages related skills with \emph{shared assets}: convention files, templates, and reference documents loaded by multiple skills in the set, keeping their outputs mutually coherent.

\subsection{Shared Assets: Cross-Skill Behavioral Coherence}

The defining feature of a skillset is the \texttt{assets/} directory at the skillset root, distinct from each skill's per-skill \texttt{assets/}. Files placed there are referenced by relative path from each constituent skill's \texttt{SKILL.md}.

Skilldex ships three reference skillsets: \texttt{developer}, \texttt{research}, and \texttt{skillset-creator}. The \texttt{developer} skillset packages four developer workflow skills around a single shared \texttt{commit-conventions.md} asset:

\begin{lstlisting}[language={}, caption={developer skillset directory layout}]
developer/
  SKILLSET.md
  assets/
    commit-conventions.md   <- shared by 2 skills
  conventional-commit/
    SKILL.md                <- references ../assets/
  changelog-gen/
    SKILL.md                <- references ../assets/
    scripts/parse-git-log.sh
    assets/changelog-template.md
  pr-description/
    SKILL.md
    assets/pr-template.md
  test-writer/
    SKILL.md
    references/testing-patterns.md
    scripts/detect-framework.sh
\end{lstlisting}

The \texttt{conventional-commit} skill writes commit messages using commit type definitions from the shared asset; \texttt{changelog-gen} classifies the same commits into changelog sections using the identical type-to-section mapping from the same file. A commit written by one skill is guaranteed to parse in the other because both are bound to the same vocabulary at install time.

The \texttt{research} skillset demonstrates the same pattern with a different artifact: an \texttt{audience-personas.md} file shared between \texttt{technical-explainer} and \texttt{paper-summarizer}. Both calibrate output vocabulary to named personas defined in the shared file, so a paper summary for the \texttt{senior-dev} persona and a plain-language explanation of the same topic share consistent calibration.

\subsection{SKILLSET.md Format}

A skillset is a directory with a \texttt{SKILLSET.md} root file. Its frontmatter mirrors \texttt{SKILL.md} exactly, with one addition: a \texttt{skills} list for remote skill references. Embedded skills, subdirectories containing their own \texttt{SKILL.md}, are auto-discovered and need not be listed explicitly.

\begin{lstlisting}[language=YAML, caption={developer SKILLSET.md frontmatter}]
---
name: developer
description: "Skills for everyday developer workflows.
  Covers commit message writing using the Conventional
  Commits spec, PR description generation, changelog
  production from git history, and test writing that
  matches a project's existing patterns."
version: "1.0.0"
tags: [developer, git, testing, workflow, productivity]
author: "skilldex-examples"
spec_version: "1.0"
---
\end{lstlisting}

\subsection{Skillset Validation Scoring}

The skillset validator performs seven checks:

\begin{table*}[t]
\centering
\caption{Skillset format conformance checks}
\label{tab:skillset-checks}
\begin{tabular}{lrp{9cm}}
\toprule
\textbf{Check} & \textbf{Pts} & \textbf{Rationale} \\
\midrule
YAML frontmatter parseable & 25 & Fatal if missing \\
\texttt{name} field present & 10 & Required for registry \\
\texttt{description} present & 10 & Discovery signal \\
Description $\geq 30$ words & 10 & Sufficient specificity \\
$\geq 1$ skill present & 20 & Empty skillset is vacuous \\
No unknown top-level dirs & 10 & Structure conformance \\
Remote URLs are GitHub URLs & 15 & Validates \texttt{source\_url} fields \\
\bottomrule
\end{tabular}
\end{table*}

\subsection{Installation Semantics}

Installing a skillset is an orchestration operation, not a new primitive. The skillset installer delegates all actual skill installation to the existing \texttt{installFromPath} and \texttt{installFromGitUrl} functions. Its algorithm is:

\begin{enumerate}
  \item Run \texttt{validateSkillset()}; abort on errors.
  \item Discover embedded skill subdirectories (those containing \texttt{SKILL.md}).
  \item For each embedded skill: call \texttt{installFromPath(embeddedSkillDir, options)}.
  \item For each remote skill reference: call \texttt{installFromGitUrl(ref.source\_url, ...)}.
  \item Copy \texttt{SKILLSET.md} and \texttt{assets/} to \texttt{skillsetsDir/<name>/}.
  \item Record the skillset in the manifest, listing embedded and remote skill names.
\end{enumerate}

Steps 3 and 4 reuse the full existing install stack, including validation, scope conflict detection, and manifest recording. The manifest entry for an installed skillset is:

\begin{lstlisting}[language=TypeScript, caption={InstalledSkillset manifest entry}]
interface InstalledSkillset {
  name: string
  version: string
  source: 'official' | 'community' | 'local'
  sourceUrl?: string
  installedAt: string
  specVersion: string
  score: number
  path: string             // relative: "skillsets/<name>"
  embeddedSkills: string[] // names of embedded skills
  remoteSkills: string[]   // names of remote skills
}
\end{lstlisting}

\subsection{Backward Compatibility}

The \texttt{skillsets} field in the manifest schema uses Zod's \texttt{.default({})} transform, so all existing \texttt{skilldex.json} manifests written before skillset support was added parse cleanly without migration.

\section{MCP Integration}
\label{sec:mcp}

Skilldex exposes a complete MCP server (\texttt{skillpm mcp}) that runs as a long-lived stdio process and can be registered in any MCP-compatible agent's configuration. All core operations are available as MCP tools:

\begin{table*}[t]
\centering
\caption{MCP tool surface}
\label{tab:mcp}
\begin{tabular}{lp{9cm}}
\toprule
\textbf{Tool} & \textbf{Description} \\
\midrule
\texttt{skilldex\_install}   & Install skill from path or \texttt{git+https://} URL \\
\texttt{skilldex\_uninstall} & Remove skill from a scope \\
\texttt{skilldex\_validate}  & Validate skill folder, return score and diagnostics \\
\texttt{skilldex\_list}      & List installed skills across scopes \\
\texttt{skilldex\_search}    & Search registry by query, tier, and limit \\
\texttt{skilldex\_suggest}   & Generate skill proposals for project \\
\midrule
\texttt{skilldex\_skillset\_install}   & Install skillset from path, registry name, or \texttt{git+https://} URL \\
\texttt{skilldex\_skillset\_uninstall} & Remove skillset and its skills from a scope \\
\texttt{skilldex\_skillset\_list}      & List installed skillsets across scopes \\
\texttt{skilldex\_skillset\_validate}  & Validate skillset folder, return score and diagnostics \\
\bottomrule
\end{tabular}
\end{table*}

All tools accept and return structured JSON. The MCP server is intentionally thin: every tool dispatches to the same \texttt{core/} functions the CLI uses, so the two interfaces cannot diverge.

This addresses a friction point in the suggestion loop (Section~\ref{sec:suggestion}): with the MCP server registered, an agent can propose, request approval, and install --- all within a single session.

\section{Implementation Details}
\label{sec:implementation}

\subsection{CLI Command Structure}

The CLI is built with Commander and follows a consistent two-file pattern for each command: a thin registration file (\texttt{command.ts}) that defines the Commander subcommand, and an action file (\texttt{command-action.ts}) that implements the logic. The registration file lazy-imports the action file, ensuring startup cost is proportional to the command invoked.

Top-level commands: \texttt{install}, \texttt{uninstall}, \texttt{update}, \texttt{list}, \texttt{validate}, \texttt{publish}, \texttt{search}, \texttt{suggest}, plus \texttt{config} (a parent command with \texttt{get}/\texttt{set}/\texttt{unset}/\texttt{list} subcommands for managing the user's local Skilldex configuration). A hidden \texttt{mcp} subcommand starts the MCP server. The \texttt{skillset} command is a parent command with seven subcommands: \texttt{init}, \texttt{install}, \texttt{publish}, \texttt{list}, \texttt{validate}, \texttt{uninstall}, \texttt{update}.

\subsection{Manifest Atomicity}

The manifest is written by serializing to JSON, writing to a temporary file, then renaming. The current implementation uses sequential writes for simplicity; concurrent writes from multiple CLI processes could produce a torn manifest. This is acceptable given the CLI's interactive use pattern.

\subsection{Git URL Parsing}

The git URL parser supports the GitHub tree URL syntax commonly used when linking to a subdirectory of a monorepo:
\begin{center}
\url{git+https://github.com/user/repo/tree/branch/subpath}
\end{center}
The parser extracts repository URL, branch, and subpath. The installer clones with \texttt{--depth 1} to the specified branch, then searches from the resolved subpath for skill directories.

\subsection{Registry Backend}

The registry is a Hono application deployed to Vercel Edge Functions. Hono was chosen for its minimal overhead on edge runtimes and its first-class TypeScript support. Supabase provides PostgreSQL (with \texttt{pg\_trgm} enabled), GitHub OAuth, and row-level security policies.

\subsection{Web Frontend}

The Skilldex web frontend is a Next.js 14 application deployed to Vercel at \texttt{skilldex-web.vercel.app}. It serves two distinct purposes within a single deployment.

\textbf{Registry browser.} The \texttt{/registry} route provides a searchable, filterable view of all published skills and skillsets, with separate tabs for each. Each entry links to a detail page (\texttt{/registry/:name} and \texttt{/registry/skillsets/:name}) showing full metadata, conformance score, trust tier badge, install count, and a one-line install command the user can copy directly.

\textbf{Documentation site.} The \texttt{/docs} route renders full documentation organized into four sections: \textit{Getting Started}, \textit{Concepts}, \textit{CLI Reference} (one page per command), and \textit{Publishing}. A dedicated \texttt{/install} page provides platform-specific installation instructions for all four supported distribution methods (npm, Homebrew, Scoop, curl) with copy-ready commands for each.

\subsection{Technology Stack}
Table~\ref{tab:stack} summarizes the libraries and platforms used across the CLI, registry, and web frontend.

\begin{table}[h]
\centering
\caption{Technology stack}
\label{tab:stack}
\footnotesize
\begin{tabular}{@{}ll@{}}
\toprule
\textbf{Component} & \textbf{Technology} \\
\midrule
CLI runtime        & Node.js 20+, TypeScript (ESM) \\
CLI framework      & Commander 13 \\
Git operations     & simple-git 3 \\
MCP server         & \texttt{@modelcontextprotocol/sdk} 1.10 \\
Schema validation  & Zod 3.24 \\
AI suggestions     & \texttt{@anthropic-ai/sdk} 0.51 \\
Registry framework & Hono 4.7 \\
Registry database  & Supabase (PostgreSQL 15) \\
Registry auth      & Supabase Auth (GitHub OAuth) \\
Rate limiting      & In-memory (Upstash Redis planned) \\
Web UI             & Next.js 14, Tailwind CSS \\
Deployment         & Vercel (registry + web) \\
Distribution       & npm, Homebrew, Scoop, curl \\
\bottomrule
\end{tabular}
\end{table}

\section{Design Philosophy}
\label{sec:philosophy}

Three principles shape Skilldex's interface choices.

\textbf{Warnings over blockers.} Validation, trust-tier checking, and cross-scope conflict detection inform but never prevent. A 12/100 skill can be installed; a community-tier skill can be installed; a global-scope installation can be overwritten at project scope with \texttt{--force}. Over-restrictive gatekeeping drives users toward workarounds, and the informational value of quality signals degrades when users learn to ignore them as installation blockers~\cite{cappos2008look}.

\textbf{Spec ownership stays with Anthropic.} Skilldex tracks and validates against Anthropic's published skill format, but does not extend, interpret, or modify it. This prevents Skilldex from becoming a de facto spec authority in tension with Anthropic's role as the canonical voice on what a well-formed skill is.

\textbf{Two interfaces, one core.} The CLI targets humans (rich terminal output, interactive prompts, colored diagnostics); the MCP server targets agents (JSON in, JSON out, no prompts). Both invoke the same \texttt{core/} modules, so the two interfaces cannot diverge behaviorally.

\section{Limitations and Future Work}
\label{sec:limitations}

We identify seven open problems acknowledged but not resolved in the current design.

\textbf{Spec versioning.} Anthropic does not currently publish the skill format as an explicitly versioned artifact. Skilldex tracks the \texttt{SKILL.md} creator guide for changes manually.

\textbf{Description quality scoring.} Format conformance can be scored objectively: word count, parseable YAML, allowed subdirectories. Semantic description quality, whether a description is specific enough to trigger reliably in context, cannot. A 30-word description that uses the right terminology triggers better than a 100-word description of the wrong thing. Possible future approaches include embedding-based similarity scoring against successful trigger contexts.

\textbf{Manifest concurrency.} The manifest write is not atomic under concurrent invocations. This would need addressing for scripted or parallel CI environments.

\textbf{Hierarchy depth.} Three levels (global/shared/project) is a design assumption. Teams with complex multi-project structures may want arbitrary nesting.

\textbf{Team governance.} The current model has no concept of team identity or role-based access, which is a gap for organizational deployments.

\textbf{Cold start on community contributions.} The registry's value scales with the number of published skills. Seeding with Anthropic's verified skills solves day-one emptiness, but community growth depends on adoption: low utility $\rightarrow$ low adoption $\rightarrow$ few contributors $\rightarrow$ low utility.

\textbf{Suggestion loop does not auto-install registry skills.} Phase 3 of the suggestion loop presents approved registry-available skills but does not invoke the installer; the user must run \texttt{skillpm install <name>} manually for each approved skill. Direct installation from the suggestion loop is planned.

\section{Conclusion}
\label{sec:conclusion}

We presented Skilldex, a package manager and registry for agent skill packages. Community install tooling already exists; Skilldex closes two gaps the existing tools leave open. Compiler-style format conformance scoring gives publishers actionable, line-level feedback grounded in Anthropic's skill specification rather than social proxies. The skillset abstraction lets related skills share common assets --- vocabulary files, templates, persona definitions --- so their outputs remain mutually coherent in a way independent installation cannot guarantee. A complete developer or research agent configuration becomes a first-class publishable artifact.

The supporting infrastructure of hierarchical scope, suggestion loop, MCP server, and metadata-only registry makes these contributions practical to use. Skilldex is open-source at {\small\url{https://github.com/Pandemonium-Research/Skilldex}}; the CLI is published to npm as \texttt{skilldex-cli}.

\section*{Acknowledgment}

The authors thank the Anthropic team for publishing the Claude skill format specification and skills directory, which provided both the foundational format that Skilldex validates and the initial corpus of verified skills that seed the registry.

\bibliographystyle{IEEEtran}

\end{document}